\documentclass[sigconf]{acmart}

\AtBeginDocument{%
  }
\usepackage{algpseudocode}
\usepackage[linesnumbered,ruled]{algorithm2e}
\usepackage{colortbl}
\copyrightyear{2022}
\acmYear{2022}
\setcopyright{acmcopyright}\acmConference[MM '22]{Proceedings of the 30th ACM
International Conference on Multimedia}{October 10--14, 2022}{Lisboa, Portugal}
\acmBooktitle{Proceedings of the 30th ACM International Conference on Multimedia
(MM '22), October 10--14, 2022, Lisboa, Portugal}
\acmPrice{15.00}
\acmDOI{10.1145/3503161.3547840}
\acmISBN{978-1-4503-9203-7/22/10}

\begin{document}

\title{Efficient Modeling of Future Context for Image Captioning}

\author{Zhengcong Fei, , Junshi Huang*, Xiaoming Wei, Xiaolin Wei}
\thanks{*Corresponding author.}
\affiliation{%
%	\institution{$^1$Key Laboratory of Intelligent Information Processing, Institute of Computing Technology, CAS, Beijing, China}
%	\institution{$^2$School of Computer Science and Technology, University of Chinese Academy of Sciences, Beijing, China}
%	\institution{$^3$School of Information Science and Technology, University of Science and Technology of China, Hefei, China}
	\institution{Meituan}
	 \city{Beijing}
	 \country{China}
}
\email{feizhengcong@meituan.com}

\renewcommand{\shortauthors}{Zhengcong Fei}
%% If needed use a foot or author note to identify equal contribution
\begin{abstract}
  Existing approaches to image captioning usually generate the sentence word-by-word from left to right, with the constraint of conditioned on local context including the given image and history generated words. There have been many studies target to make use of global information during decoding, e.g., iterative refinement. However, it is still under-explored how to effectively and efficiently incorporate the future context. To respond to this issue, inspired by that Non-Autoregressive Image Captioning (NAIC) can leverage two-side relation with modified mask operation, we aim to graft this advance to the conventional Autoregressive Image Captioning (AIC) model while maintaining the inference efficiency without extra time cost. Specifically, AIC and NAIC models are first trained combined with shared visual encoders, forcing the visual encoder to contain sufficient and valid future context; then the AIC model is encouraged to capture the causal dynamics of cross-layer interchanging from NAIC model on its unconfident words, which follows a teacher-student paradigm and optimized with the distribution calibration training objective. Empirical evidences demonstrate that our proposed approach clearly surpass the state-of-the-art baselines in both automatic metrics and human evaluations on the MS COCO benchmark. The source code is available at: https://github.com/feizc/Future-Caption.
\end{abstract}

\begin{CCSXML}
<ccs2012>
<concept>
<concept_id>10010147.10010178.10010224</concept_id>
<concept_desc>Computing methodologies~Computer vision</concept_desc>
<concept_significance>500</concept_significance>
</concept>
<concept>
<concept_id>10010147.10010178.10010179</concept_id>
<concept_desc>Computing methodologies~Natural language processing</concept_desc>
<concept_significance>300</concept_significance>
</concept>
</ccs2012>
\end{CCSXML}
\ccsdesc[500]{Computing methodologies~Computer vision}
\ccsdesc[300]{Computing methodologies~Natural language processing}
\keywords{Future Context; Image Captioning; Non-autoregressive Decoding; Causal Dynamics Calibration}


\maketitle

\section{Introduction}

Image captioning, which aims to describe the image content with natural language, has seen rapid development in the past several years \cite{chen2015microsoft}. 
In a conventional image captioning system, an visual encoder first transform the given image into a sequence of intermediate hidden representations, based on which, a language decoder generate the sentence word by word. Such encoder-decoder paradigm is usually implemented by CNN-LSTM \cite{Vinyals2015Show,Xu2015Show} or Transformer \cite{attention} network architecture, and optimized with teacher forcing objectives \cite{Anderson2017Bottom,herdade2019image,huang2019attention,cornia2020meshed,pan2020x,zhang2021rstnet}. 
Despite its success, the autoregressive structure of the left-to-right manner makes the model only access to the local context, \emph{i.e.}, the previously generated words as well as the given image, at each decoding step. Such a unidirectional property makes the models unable to exploit global context effectively, yielding an unsatisfied description. \cite{wang2016image,wang2018image,zhou2022compact}.

% 缺少对图1的ref

\begin{figure}
	\begin{center}
	\includegraphics[width=0.9\columnwidth]{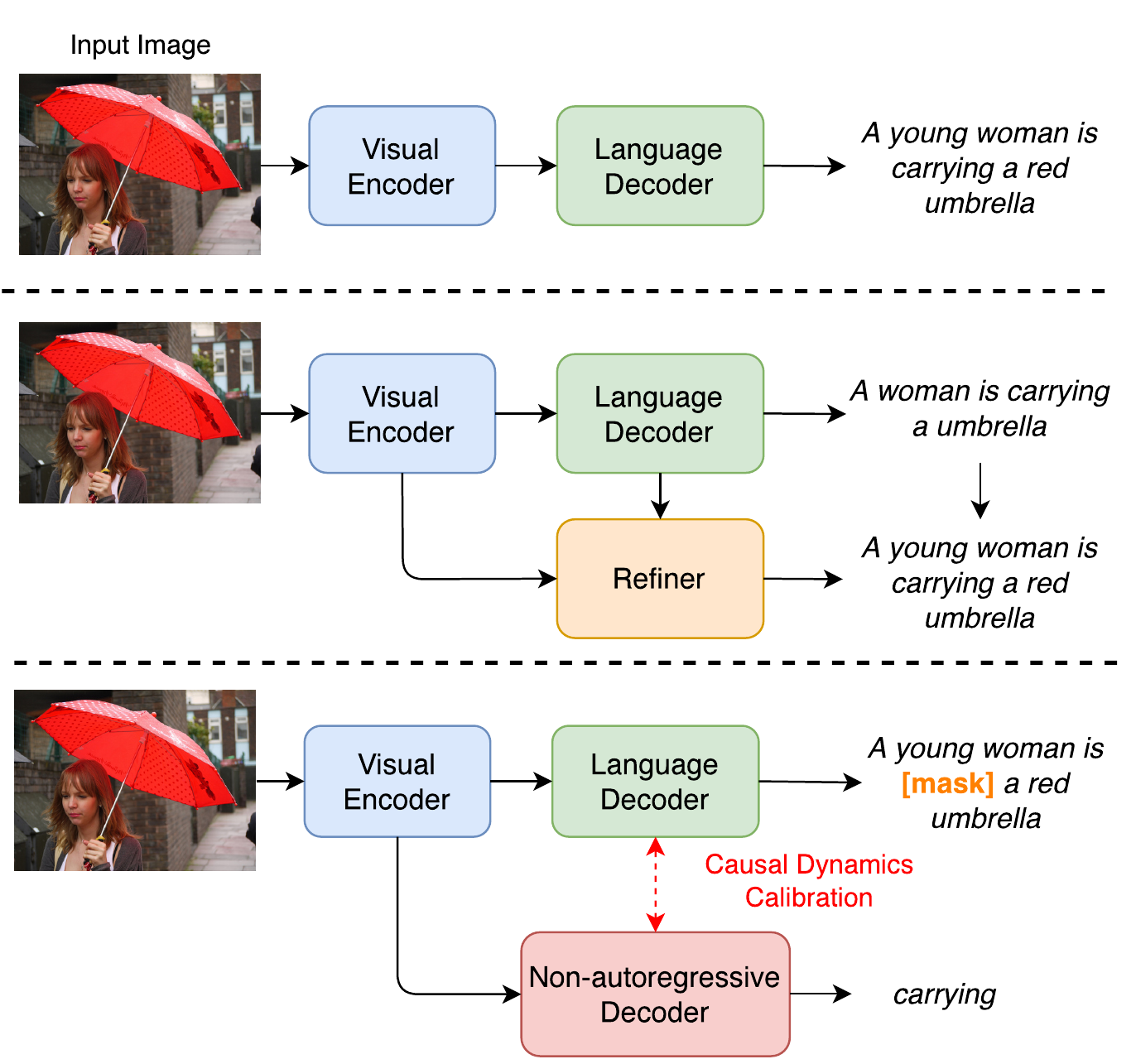}
	\end{center}
	\caption{Overview of conventional image captioning, refinement-based image captioning, and our future context modeling with causal dynamics calibration from non-autoregressive decoder. Note that the non-autoregressive decoder is not involved at the inference stage to maintain computation efficiency. }
	\label{fig:1}
\end{figure}

To address this issue, many researchers have attempted to exploit the global information during sentence generation. Typically, refinement-based method is introduced \cite{sammani2019look,khademi2018image,sammani2020show,wang2020show,song2021image,zhang2021vinvl,yan2021semi,wang2019neural}, which typically consists of two networks: the first network is usually a primary generator or an image-text retrieval, which is used to generate or retrieve a coarse related template; A refiner is in series to generate the final caption by attending to the sentence produced before. Such an iterative refinement operation can help the model to look at both past and future semantic context and thus improve decoding at every time step. However, most of the works rely on multi-pass decoding or specially customized decoding algorithms which leads to a significant increase in training and inference costs.
On the other hand, modeling the global context in the reverse direction by pairing the conventional left-to-right image captioning model with a right-to-left auxiliary model is also delivered \cite{wang2018image,sammani2020show,stefanini2021show,duan2021modeling}. However, in these methods, the modeling of reverse context is still conditioned on the local context with a separate network and they cannot sufficiently encourage the image captioning model to exploit a truly flexible global context.

In pursuit of effectively and efficiently incorporating global information into image captioning models, we conduct well-designed pilot experiments, and find some interesting phenomenons: i) Even conditioned on absolutely correct context, i.e., historical words and given image, there is a certain proportion of ground-truth words that the image captioning model predicts with relatively low probabilities. ii) The probability assigned to the ground-truth words from image captioning models differs with the captioning length. In contrast, with the factorized probability modeling, a good image captioning model should endow the highest probability to correct words according to accurate historical information. 
Consistent with \cite{zhou2022compact,zhou2019synchronous,gu2018stack}, we believe that the reasonable cause of this phenomenon is that the image captioning model cannot confidently predict these words according to only the local context. Therefore, it should be improved for the image captioning model on these unconfident words with sufficient distribution calibration.

In this paper, we introduce efficient modeling of future context information for image captioning, referred to as FutureCap. 
In general, the architecture of the original autoregressive image captioning (AIC) model is kept untouched and jointly optimized with an additional mask-based non-autoregressive image captioning (NAIC) model \cite{gao2019masked,guo2020non,fei2019fast}, which is essentially cross-modal understanding and contains the global context. As shown in Figure \ref{fig:1},  the AIC model and NAIC model are first trained combined in multi-task learning with sharing the visual encoders.  The visual encoder is additionally supervised by the signal from the NAIC decoder to include sufficient future context information; Then, we employ casual dynamics calibration that pushes the student AIC model faithfully learn the causal effect of teacher NAIC model representations on unconfident output, with cross-layer interchanging aligning. This further help the AIC model leverage knowledge to information dynamics.  
Experimentally, we evaluate our approach to the MS COCO dataset. According to both automatic metrics and human evaluations, the captioning models equipped with future context modeling evidently outperform baselines. 
The major \textbf{contributions} of our paper are as follows:
\begin{itemize} 
    \item We focus on the efficient modeling of future information for better image caption decoding and analyze the necessity of global context clearly with pilot experiments. 
    \item We introduce causal dynamics calibration that encourages the student AIC model to learn the interchange aligning from teacher NAIC model on unconfident words and adjust knowledge routing with share visual encoder, to more effectively exploit the future contextual information.
    \item  Experiments on the MS COCO dataset demonstrate that image captioning models equipped with our future context modeling framework significantly outperforms the one without it. More encouragingly, as most of the previous literature improves the performance by increasing the model capacity, our approach represents a new optimization paradigm that leads to no additional inference cost.
\end{itemize}

\section{Background and Pilot Analysis}
To investigate the potential impact of future context in image captioning, we first describe the basic architectures of conventional autoregressive and non-autoregressive image caption models, both of which follows on Transformer-based encoder-decoder paradigm.  After that, we conduct pilot experiments as well as empirical analyses on the effects of context information for caption decoding.

\subsection{Model Architecture}

Generally, AIC and NAIC models hold the same visual encoder architecture while differing in decoders for their mask matrix of self-attention mechanisms and prediction manners.

\paragraph{\textbf{Visual Encoder}}
The visual encoder aims to learn the high-level visual representations of the given image, which includes $L$ same network layers. Each layer consists of two sub-layers: a self-attention sub-layer and a position-wise feed-forward network sub-layer. The input of the network layer is the hidden states of the previous layer, on which the multi-head scaled dot-product attention computation is performed. Assuming that $h^l_e$ presents the hidden states of the $l$-th encoder layer, the visual encoder layer can be computed as: 
\begin{align}
    s^l = \text{SelfAttention}(h^{l-1}_e, h^{l-1}_e, h^{l-1}_e), \\
    h^l_e = \text{FeedForward}(s^l).
\end{align}
Layer normalization with residual connection is added after both two sub-layers. Note that $h^0_e$ is initialized as the patch embedding of the extracted image region features, and the hidden states of the $L$-th layer $h^L_e$ are served as input to the language decoder.

\paragraph{\textbf{Language Decoder}}

The decoder of the AIC and NAIC models are introduced separately. For the autoregressive decoder, which usually consists of three sub-layers: a masked self-attention sub-layer, a cross-attention sub-layer, and an FFN sub-layer. In particular, to maintain the autoregressive generated property at each time step, the masked self-attention sub-layer performs self-attention with a causal attention mask to prevent the decoder from seeing subsequent words. To generate the hidden states $h^l_d$ of the $l$-th decoder layer, the autoregressive decoder can be formulated as:
\begin{align}
    s^l = \text{MaskSelfAttention}(h^{l-1}_d, h^{l-1}_d, h^{l-1}_d),\\
    c^l = \text{CrossAttention}(s^l, h^L_e, h^L_e),\\
    h^l_d = \text{FeedForward}(c^l).
\end{align} 
Layer normalization with residual connection is also added after each sub-layers. Finally, with the given image $x$, the generated sentence $w_{<t}$ and the learned top-layer hidden states $h_{d,t}$, the decoder models the probability distribution as:
\begin{equation}
    p_{AIC}(w_t|w_{<t}, x) = \text{Softmax}(W h_{d,t}),
\end{equation}
where $W$ denotes the learnable parameter.

\begin{figure}
	\begin{center}
		\includegraphics[width=0.9\columnwidth]{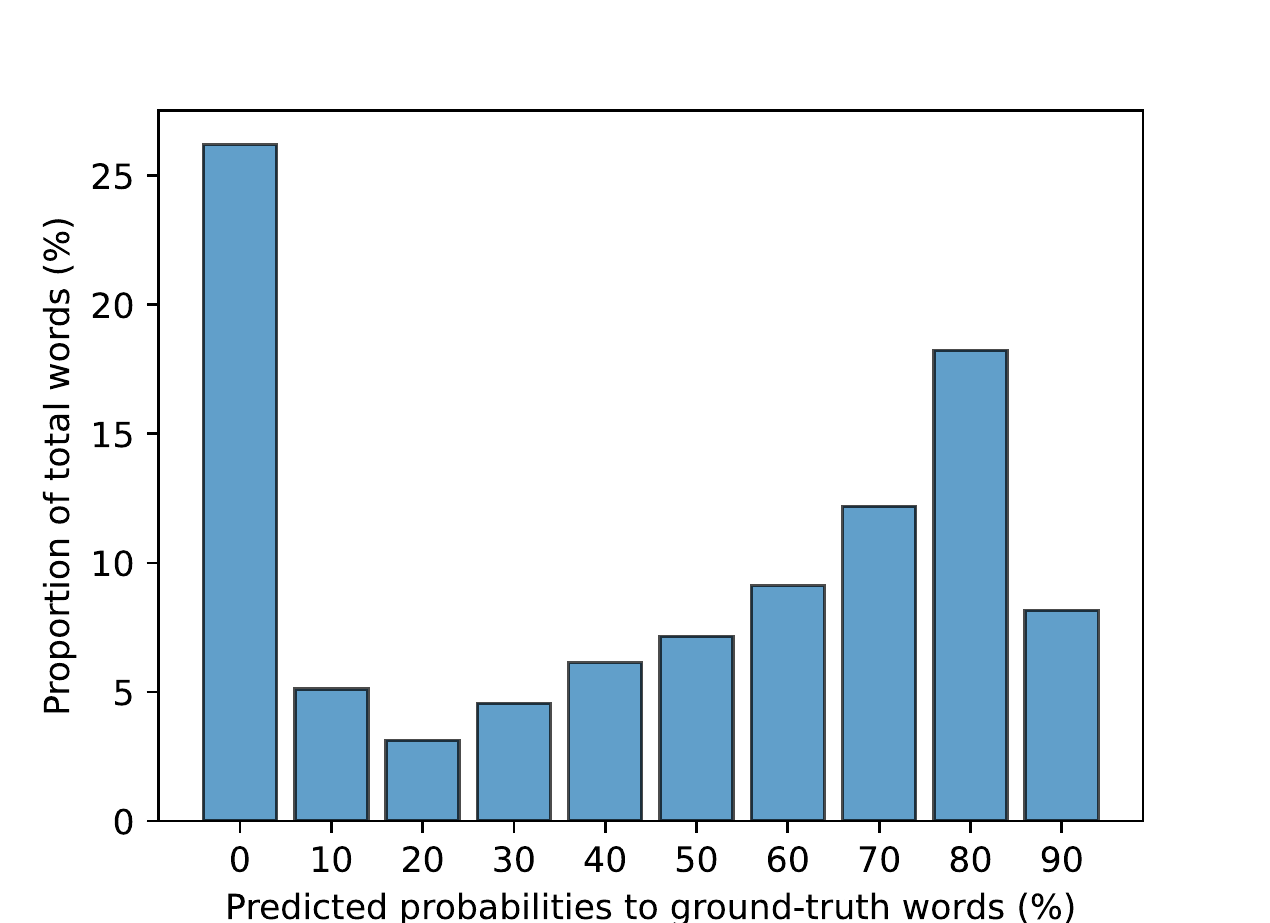}
	\end{center}
	\caption{Predicted probability of the different ground-truth words on the training set of MS COCO dataset.}
	\label{fig:2}
\end{figure}

\begin{figure}
	\begin{center}
		\includegraphics[width=0.9\columnwidth]{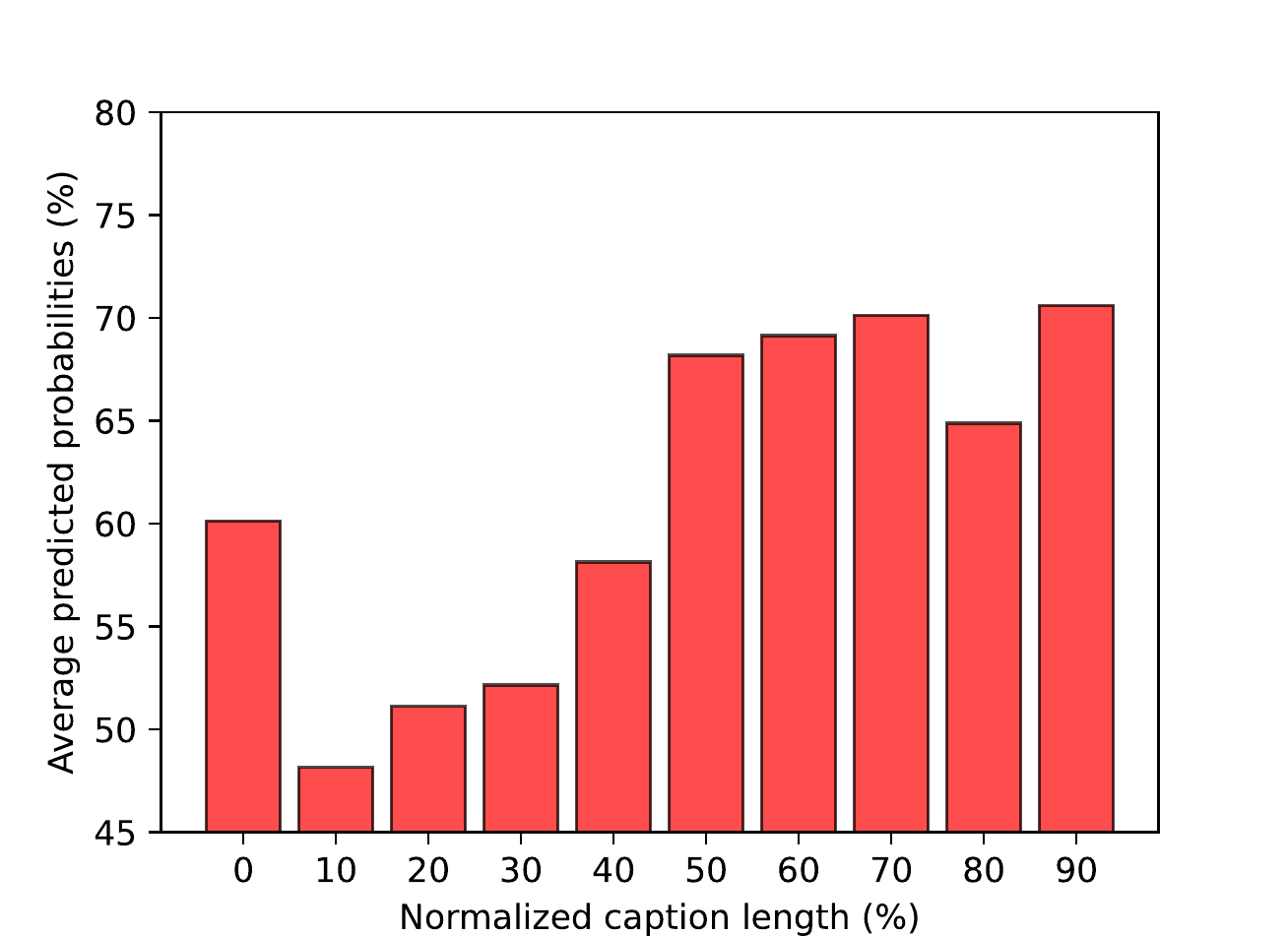}
	\end{center}
	\caption{Average predicted probability of the ground-truth words for the different normalized sentence length on the training set of MS COCO dataset.}
	\label{fig:3}
\end{figure}

For the non-autoregressive decoder, which aims to predict a set of masked target words $S_m$ given an image $x$ and a set of observable target words $S_o$. 
The NAIC decoder also contains the same $L$ identical layers, each of which also includes a self-attention sub-layer, a cross-attention sub-layer, and a feedforward sublayer. Unlike the masked self-attention sub-layer of the AIC decoder, the attention mask is removed in the NAIC decoder.
Finally, with the learned top-layer hidden states $\tilde{h}_{d}$ of the NAIC decoder and partially observed sentence $S_o$, the predicted probability distribution for every masked word $w_t \in S_m$ can be calculated as:
\begin{equation}
    p_{NAIC}(w_t|S_o, x) = \text{Softmax}(\tilde{W} \tilde{h}_{d,t}),
\end{equation}
where $\tilde{W}$ is the learnable parameter. Note that since the decoder of the NAIC model takes $S_o$ rather than $w_{<t}$ as input, which includes both history and future words with respect to every masked target word, it should embody the global contextual information.

\subsection{Are History Contexts Enough for Prediction? }

A high-quality image captioning model is supposed to endow the highest probabilities of the ground-truth words based on correct historical context. Delicate experiments are conducted in this section to explore the disadvantage of conventional transformer image captioning with a teacher-forcing training framework.

\paragraph{\textbf{Experimental Setting}}

Experimentally, we adopt the basic configuration of the Transformer-based image captioning model without the mesh-memory module, which is publicly available at GitHub \footnote{https://github.com/aimagelab/meshed-memory-transformer}. To be specific, the model is comprised of 6 standard transformer layers of visual encoder and language decoder.
Moreover, the regional features of images extracted from faster r-cnn \cite{ren2015faster} on the backbone of ResNet \cite{he2016deep} are utilized to retrain the image captioning model with the current configuration \cite{attention}. For the training process, the image captioning model is first trained with cross-entropy loss and then fine-tuned with sentence-level self-critical reward \cite{Rennie2017Self} follow the default training settings with Adam \cite{kingma2014adam} optimizer on the MS COCO \cite{chen2015microsoft} dataset of Karpathy training split.

After obtaining a fully-trained image captioning model, we record the predicted probability of the ground-truth words given the correct context including image and the previous subsentence in the MS COCO training set. 
In order to characterize the results, we plot the proportion of total words in different predicted probability chunks in Figure \ref{fig:2}. Meantime, we also plot the average predicted probability of the corresponding ground-truth words for different caption lengths in Figure \ref{fig:3}. Here we normalized the caption to eliminate the influence of absolute sentence length.

\begin{figure*}
	\begin{center}
		\includegraphics[width=1.8\columnwidth]{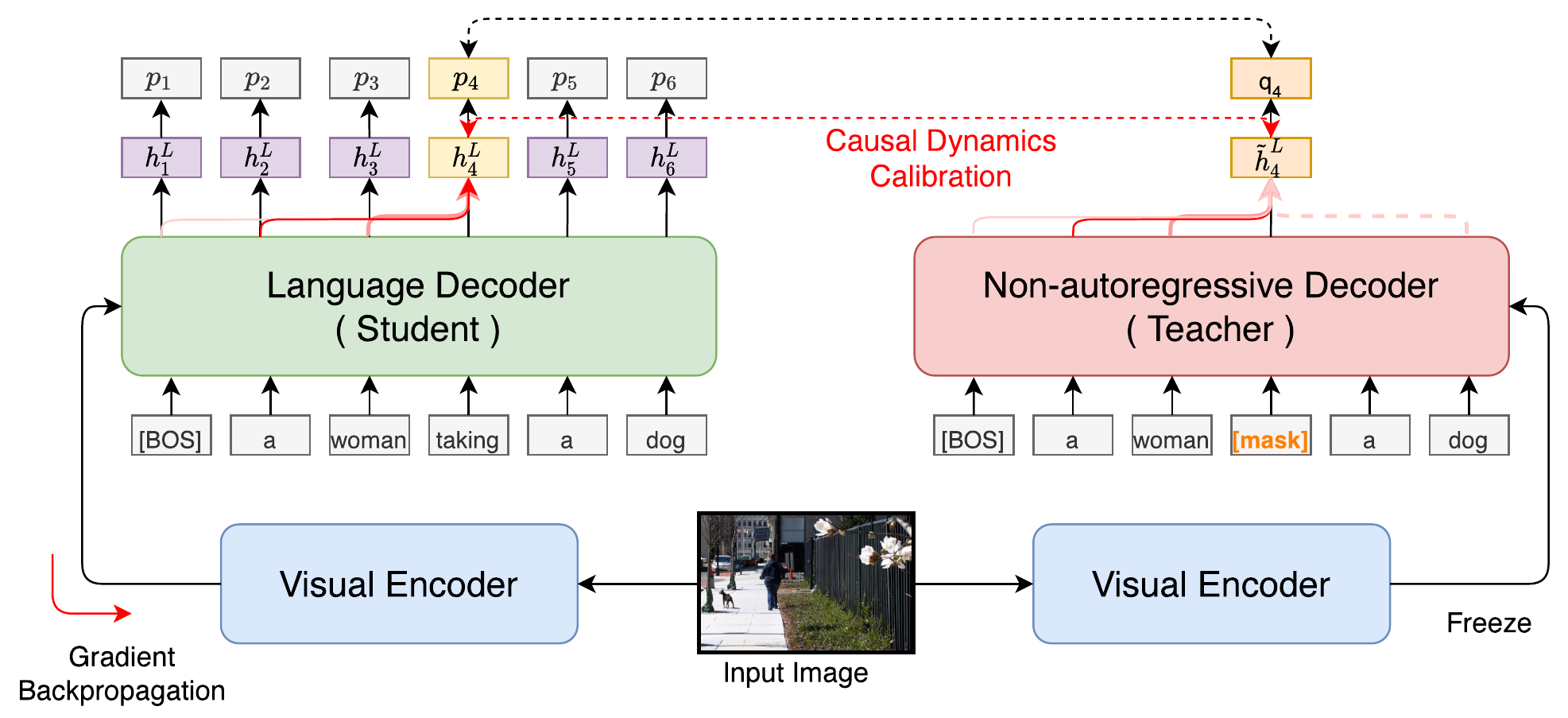}
	\end{center}
	\caption{Illustration of future context modeling with non-autoregressive decoder for language decoder.
	Assuming that the confidence of predicted word $\{p_4 \}$ from language decoder is lower than threshold $\epsilon$, the masked words set $S_m$ becomes $\{w_4\}$. Thus the input $S_o$ to the NAIC decoder is $\{w_1, w_2, w_3, [mask], w_5, w_6\}$, and the output hidden states $\tilde{h}_4^L$ and probability $q_4$ is used to calibrate the causal dynamics of original $p_4$. Note that the total parameters of the NAIC model is freezed.}
	\label{fig:4}
\end{figure*}

\paragraph{\textbf{Results Discussion}}

According to the estimation results of Figure \ref{fig:2}, we can find that even provided with the totally correct context, there is an obvious portion of ground-truth words that the conventional image captioning model predicts with relatively low probabilities.  For instance, the model predicts 25.67\% ground-truth words with probabilities chunk between 0.0-0.1. The reasonable cause of this phenomenon is that  \textbf{image captioning model cannot confidently predict these ground-truth words according to only the local context of image and history words}. 

To further prove where the low confidence words locate, we calculate the average predicted probability for various locations in the total length. We can see that with the relative length increase, \emph{i.e.}, the generation position moves from left to right, and the average probability of ground-truth words increases gradually. We attribute that with the generated sentence length increase, the determined context increase, and future context becomes less, which results in the strengthening of model confidence. 
According to the experimental results, it is natural to consider improving the image captioning model on these correct while unconfident words with effective future information to assist the current decision.

\section{Methodology}

Based on the above analysis, we believe that modeling future information in image captioning, especially for unconfident words, is necessary.
To this end, we propose a new framework for efficient modeling of future context, named FutureCap, that tries to employ a mask-based non-autoregressive image captioning decoder to enhance the conventional image captioning model according to its prediction confidence. 
Overall, we first train the AIC model and NAIC model with a shared visual encoder to acquire an image encoding supervision. Then, we employ the NAIC model as a teacher to improve the unconfident word of the student AIC model through causal dynamics calibration.

\subsection{Shared Visual Encoder Supervision}

To encourage the visual encoder containing sufficient global information, we first train the image captioning model and NAIC model with shared visual encoder in a multi-task manner and optimize the combined training objective as follows:
\begin{equation}
    {L}_{VE}(\theta_{VE}, \theta_{AIC}, \theta_{NAIC}) = \lambda  {L}_{AIC} + (1-\lambda) {L}_{NAIC}, \label{eq:10}
\end{equation}
where $\theta_{VE}$, $\theta_{AIC}$ and $\theta_{NAIC}$ denote the parameters of the shared visual encoder, the autoregressive language decoder, and the mask-based non-autoregressive decoder, respectively. 
$\lambda$ is a balancing factor between two losses. 
As the visual encoder is additionally supervised by the signal from the mask-based NAIC decoder, the AIC model is able to disentangle the future information from the extracted visual representation.
In between, the AIC model is first optimized through the time-wise cross-entropy loss:
\begin{equation}
    {L}_{AIC}(\theta_{VE}, \theta_{AIC}) = -\sum_{t=1}^{|S|} \text{log }p(w_t|w_{<t}, I).
\end{equation}
And then fine-tuning using with CIDEr score reward $r$ and mean baseline $b$. The gradient expression for SCST \cite{Rennie2017Self} training is, 
\begin{equation}
\small
    \nabla_{(\theta_{VE}, \theta_{AIC})} L_{AIC} = - ((r(w^i)-b)\nabla_{(\theta_{VE}, \theta_{AIC})} \text{log}p_{AIC}(w^i)).
\end{equation}
For NAIC decoder, we adopt the strategy in \cite{gao2019masked}. Concretely, we randomly select $n$ words, and replace each selected word with a special symbol $[mask]$, splitting sentence $S$ into observed set $S_o$ and masked set $S_m$. We eventually minimize the following training objective for each word in masked sentence $S_m$ as: 
\begin{equation}
     {L}_{NAIC}(\theta_{VE}, \theta_{NAIC}) = -\sum_{w_t \in S_m} \text{log } p(w_t|S_o, I).
\end{equation}

\subsection{Causal Dynamics Calibration}

We then introduce to use NAIC model as a teacher to transfer the knowledge for the student NAIC model on unconfident words decision, i.e., help the NAIC model to capture and consider more global information from visual representation and generated words. The parameters of the teacher NAIC model are frozen in this stage.   
Figure \ref{fig:4} depicts the training procedure of this stage with an easy understanding example. Formally, given the image $I$ and the generated ground-truth words $w_{<t}$ at each time step $t$, we first ask the AIC model make predictions for every word using Equation 6, generating the prior word-level probability distributions $\{{p}_1, {p}_2, \ldots,  {p}_{|S|} \}$ and $|S|$ is the sentence length. Then, the masked word set $S_m$ is built,  where the predicted probabilities ${p}_t$ to the corresponding ground-truth words are lower than a threshold value $\epsilon$ as:
$S_m = \{w_t | {p}_t \leq \epsilon, 1 \leq t \leq |S| \}$. Next, we obtain the observed set $S_o$ for NAIC model input by replacing those selected low-confident ground-truth words in the original sentence $S$ with a special symbol $[mask]$. Note that the equation $S = S_o \cup S_m$ is always true in our framework.
Next, we can obtain the predicted probability distribution $\hat{q}_t$ from the teacher NAIC model for every word in $S_m$ using Equation 7.

\begin{algorithm}[t] 
	\label{a:1}
	\caption{Causal Dynamics Calibration Algorithm between student AIC model and teacher NAIC model}  
	\KwIn{Unconfident masked data $S_m$, student AIC model $p_{AIC}$ with output neural $N$, teacher NAIC model $p_{NAIC}$, neural alignment $g$}  
	% \KwOut{Semi-Autoregressive Image Captioning model $P_{SAIC}$}  
	Fix the parameters of $p_{NAIC}$\;
	\While{not converged}{
	\For{$w_t$ in $S_m$}  
	{   
	    $N_S$ = Sample\_student\_neurons($N$)\;
	    $N_T$ = $g(N_s)$\;
	    Compute the causal dynamic difference \\
	    || \texttt{GET}($p_{AIC}, N_S, w_t$) - \texttt{GET}($p_{NAIC}, N_T, w_t)$ ||$^2_2$ \; 
	    Compute KD loss $\text{KL}({q}_t||{p}_t)$ \;
	    Compute combined loss $L_{CDC}$ \; 
	    Loss backward\;
	    Step optimizer\;
	} 
	}
\end{algorithm}

Once get the knowledge routing of the AIC and NAIC model, to improve the decision on the set $S_m$ of its unconfident words, we introduce \emph{causal dynamics calibration} to assist the future context modeling of AIC model. The detailed progress is shown in Algorithm 1. In between, \texttt{GET} operation is defined as an activation-value retriever for a neural model. Given a model $p$ contain a set of neurons $N$, i.e., internal representations, and input context $c_t$ including image $x$ and generated word $w_{<t}$, $\texttt{GET}(p, N, c_t)$ is the set of weight values that neural $N$ takes on when processing the context $c_t$. 
For context $c_t$, $N_s$ is the set of neuros from student AIC model, we can get the interchanging alignment loss as: 
\begin{equation}
\begin{split}
    L_{IA}(\theta_{VE}, \theta_{AIC}) = \sum_{w_t \in S_m} & || \texttt{GET}(p_{AIC}, N, w_t) \\ &- \texttt{GET}(p_{NAIC}, g(N), w_t) ||^2_2, 
\end{split}
\end{equation}
where $g(\cdot)$ is the mapping function for sampled neurons except the future neurons in teacher NAIC model. Similar to the conventional knowledge distillation \cite{hinton2015distilling}, we also restricted the output distribution for unconfident words with KL-divergence as:
\begin{equation}
    {L}_{KL}(\theta_{VE}, \theta_{AIC}) = \sum_{w_t \in S_m}  \text{KL}({q}_t||{p}_t),
\end{equation}
The final training objective for the student AIC model is a combination of the three terms reviewed above as: 
\begin{equation}
\begin{split}
    {L}_{CDC}(\theta_{VE}, \theta_{AIC}) =& {L}_{KL}(\theta_{VE}, \theta_{AIC}) +  L_{IA}(\theta_{VE}, \theta_{AIC}) \\ &- \sum_{w_t \in S_o } \text{log } {p}_t,
\end{split}
\end{equation}
where the last term makes the AIC model stable to the high confidence ground-truth words. By doing so, we can fully strengthen the ability of the AIC model to leverage the global context contained in the NAIC. On the other hand, 
to avoid making the student model rely heavily on the decision of the teacher NAIC model, we also employ a teacher annealing strategy to linearly decrease the knowledge distillation to ground-truth supervision of sentence-level reward \cite{Rennie2017Self} throughout training.
Note that the NAIC model is not involved at the inference stage to keep efficiency.

\section{Experiments}

\subsection{Experimental Preparation}

\paragraph{\textbf{Dataset}}
We evaluate the proposed method on the MS COCO \cite{chen2015microsoft}, which is a standard estimation benchmark for image captioning tasks.  To be consistent with previous work \cite{huang2019attention,cornia2020meshed}, we adopted the Karpathy split \cite{karpathy2015deep} that contains 113,287 training images and 5,000 images for validation and test splits, respectively. Each image corresponds with 5 different captions. We omit words that occur less than 5 times and the vocabulary size is 10,369 words. Image features are extracted with CLIP \cite{Anderson2017Bottom} for 512-dim vectors. 

\paragraph{\textbf{Evaluation Metrics}}
Following the common paradigm, we utilize five metrics to comprehensively estimate the captioning performance: BLEU-$N$ \cite{Papineni2002BLEU}, METEOR \cite{Lavie2007METEOR}, ROUGE \cite{Flick2004ROUGE}, CIDEr \cite{dis1}, and SPICE \cite{spice}. We denote as B-$N$, M, R, C, and S for simplicity. 
% Particularly, SPICE focuses on semantic analysis and has a higher correlation with human judgment, and other metrics favor frequent $n$-grams and measure the overall sentence fluency. 

\paragraph{\textbf{Implementation Details}}
Our implementation is based on Pytorch and repository  \cite{cornia2020meshed} to build the model under Transformer-base configuration with the memory module, where AIC and NAIC hold the identical architecture. Concretely, both the network is comprised of 6 visual encoder and 6 language decoder layers, each with 512 as hidden size, the FFN sublayers of 2,048 dimensions, and 8 heads in multi-head attentions. We set the dropout rate to 0.1. The neurons are sampled from the top-layer decoder with a unifying distribution. For parameter updating, we employ the Adam optimizer \cite{kingma2014adam} with the default setting. As for learning rate schedule, we adopt the same strategy as \cite{attention, cornia2020meshed} and set warm-up steps to 4,000. In the first stage, we train all models by sharing their encoders for 300k steps. In the second stage, we separate their encoders and fix the parameter of the NAIC model. Then, the AIC model is solely optimized with the fixed NAIC by additional 200k steps.

\begin{table}[t]
	\begin{center}
	{\caption{Performance comparisons of our FutureCap model and other state-of-the-art image captioning models with different evaluation metrics on the MS COCO Karpathy test set. All values are reported as a percentage (\%).  }
			\label{tab:1}}
		\setlength{\tabcolsep}{1.6mm}{
			\begin{tabular}{lcccccc}	
				\hline
				%& \multicolumn{6}{c|}{\text{Cross-Entropy Loss}} &\multicolumn{6}{c}{\text{CIDEr Score Optimization}} \\ 
				&B-1&B-4&M&R&C&S\\
					\hline 
					\hline
				%\multicolumn{13}{l}{\emph{State-of-the-art image captioning models}}\\
				%\midrule
				LSTM-A \cite{yao2017boosting}&78.6&35.5&27.3&56.8&118.3&20.8\\
				%	\multicolumn{13}{l}{\textbf{LSTM-based}} \\ \midrule
				Up-Down \cite{Anderson2017Bottom}  &79.8&36.3&27.7&56.9&120.1&21.4\\
				GCN-LSTM \cite{exploring}&80.5&38.2&28.5&58.3&127.6&22.0\\
				AoANet \cite{huang2019attention} &80.2&38.9&29.2&58.8&129.8&22.4\\ 
				$\mathcal{M}^2$ Transformer \cite{cornia2020meshed} &80.8&39.1&29.2&58.6&131.2&22.6\\ 
				X-LAN \cite{pan2020x}
				&80.8&39.5&29.5&59.2&132.0&23.4
				\\
				DPA \cite{liuprophet} &80.3 &40.5&29.6&59.2&133.4&23.3\\ 
				GET \cite{ji2021improving} &81.5&39.5&29.3&58.9&131.6&22.8\\
				DLCT \cite{luo2021dual}
				&81.4&39.8&29.5&59.1&133.8&23.0
				\\
				RSTNet \cite{zhang2021rstnet} &81.8&40.1&29.8&59.5&135.6&23.3
				\\
					\hline
				%\multicolumn{13}{l}{\emph{Our memory-augmented image captioning models}}\\ \midrule
			  \rowcolor[gray]{.9} FutureCap &\textbf{82.2}&\textbf{40.3}&\textbf{30.1}&\textbf{59.8}&\textbf{136.3}&\textbf{23.8}\\	%\multicolumn{13}{c}{\textbf{Ensemble / Fusion}}\\ \midrule
		    	\hline
		\end{tabular}}
		\end{center}
\end{table}

\begin{table*}[t]
	\begin{center}
	{\caption{Leaderboard of different image captioning models on the online MS COCO test server. }
			\label{tab:2}}
	\setlength{\tabcolsep}{2mm}{
		\begin{tabular}{lcccccccccccccc}
			\hline
			%	& \multicolumn{5}{|c|}{\textbf{Cross-Entropy Loss}} &\multicolumn{5}{|c|}{\textbf{CIDEr-D Score Optimization}} \\ \cline{2-11}
			&\multicolumn{2}{c}{BLEU-1}&\multicolumn{2}{c}{BLEU-2}&\multicolumn{2}{c}{BLEU-3}&\multicolumn{2}{c}{BLEU-4}&\multicolumn{2}{c}{METEOR}&\multicolumn{2}{c}{ROUGE}&\multicolumn{2}{c}{CIDEr}\\ \hline
			&c5&c40&c5&c40&c5&c40&c5&c40&c5&c40&c5&c40&c5&c40\\
		\hline  \hline
			Up-Down \cite{Anderson2017Bottom}  &80.2&95.2&64.1&88.8&49.1&79.4&36.9&68.5&27.6&36.7&57.1&72.4&117.9&120.5\\
			%	GCN-LSTM  &80.8&95.9&65.5&89.3&50.8&80.3&38.7&69.7&28.5&37.6&58.5&73.4&125.3&126.5\\
			AoANet \cite{huang2019attention} &81.0&95.0&65.8&89.6&51.4&81.3&39.4&71.2&29.1&38.5&58.9&74.5&126.9&129.6\\
			$\mathcal{M}^2$ Transformer \cite{cornia2020meshed} &81.6&96.0&66.4&90.8&51.8&82.7&39.7&72.8&29.4&39.0&59.2&74.8&129.3&132.1\\
			X-Transformer \cite{pan2020x} &81.9&95.7&66.9&90.5&52.4&82.5&40.3&72.4&29.6&39.2&59.5&75.0&131.1&133.5\\
			GET \cite{ji2021improving} &81.6&96.1&66.5&90.9&51.9&82.8&39.7&72.9&29.4&38.8&59.1&74.4&130.3&132.5\\
			DPA \cite{liuprophet} &81.8&96.3&66.5&91.2&51.9&83.2&39.8&73.3&29.6&39.3&59.4&75.1&130.4&133.7 \\
			RSTNet \cite{zhang2021rstnet} &82.1&96.4&67.0&91.3&52.2&83.0&40.0&73.1&29.6&39.1&59.5&74.6&131.9&134.0
			\\
		\hline
		\rowcolor[gray]{.9}	FutureCap&\textbf{82.4}&\textbf{96.7}&\textbf{67.3}&\textbf{91.8}&\textbf{52.6}&\textbf{83.8}&\textbf{40.3}&\textbf{74.0}&\textbf{29.6}&\textbf{39.2}&\textbf{59.6}&\textbf{74.9}&\textbf{132.9}&\textbf{135.3}\\  	\hline
		\end{tabular}
			}
	\end{center}
\end{table*}

\subsection{Comparison with State-of-the-Art Models}

\paragraph{\textbf{Performance on MS COCO}}
We compare the results of our FutureCap model with those of several recent image captioning models trained without large-scale vision-and-language pre-training on the offline MS COCO dataset. The evaluation results are listed in Table \ref{tab:1}. First, we can see that FutureCap surpasses the original memory-incorporated Transformer by +0.6 BLEU-4 and 7.1 CIDEr scores, respectively, verifying that modeling the future information brings a significant performance improvement. Next, it is encouraging that our proposed framework outperforms the most recent competitive models. 
As it can be observed, our proposal reaches 136.3 CIDEr points, beating almost all the compared approaches. It is encouraging that our strategy can be combined with other advance improved strategies as well as untouched the internal structures. What's more,  as most of the previous literature has boosted caption quality by increasing the model capacity, which leads to an extra burden for application devices, our approach represents an outliner in this trend and demonstrates that state-of-the-art CIDEr levels can be obtained even with a very lightweight efficient model. 

\paragraph{\textbf{Online evaluation}}

We also report the performance of our method on the online MS COCO test server. In this case, we employ an ensemble of four models trained with the same configuration of NAIC decoder assistance, for which ground-truth annotations are not publicly available.
Comparison results with the top-performing approaches of the leaderboard are reported in Table \ref{tab:2}. As it can be seen, our method surpasses the current state-of-the-art model on all metrics, achieving an advancement of 1.3 CIDEr points with respect to the best performer.

\subsection{Model Analysis}

\begin{table}
	\begin{center}
	{\caption{Ablation studies on the MS COCO test set.}
				\label{tab:3}}
		\setlength{\tabcolsep}{1.8mm}{
			\begin{tabular}{lcccccc}
				\hline
				%	& \multicolumn{5}{|c|}{\textbf{Cross-Entropy Loss}} &\multicolumn{5}{|c|}{\textbf{CIDEr-D Score Optimization}} \\ \cline{2-11}
				Model&B-1 &B-4&M&R&C&S\\
				\hline 
				\hline
				\rowcolor[gray]{.9} FutureCap&\textbf{82.2}&\textbf{40.3}&\textbf{30.1}&\textbf{59.8}&\textbf{136.3}&\textbf{23.8}\\ 
				\emph{w/o.} VES &81.6&40.0&29.5&59.5&135.2&23.6\\
				\emph{w/o.} CDC&81.4&39.9&29.4&59.4&134.8&23.5\\ 
				\emph{w/.} KL &82.1 & 40.1&29.9&59.6&135.9&23.6\\
				\hline
			\end{tabular}
	}
	\end{center}
\end{table}

\paragraph{\textbf{Ablation Study}}

To better understand the influence of each design in our FutureCap model, we conduct ablation studies on the offline MS COCO dataset. Table \ref{tab:3} reports the evaluation results on the testing set. We first validate the necessity of shared visual encoder supervision by training the AIC and NAIC model using the separate visual encoder, respectively, denoted as  “w/o. VES”. The automatic metrics of final performance decrease by a 1.1 CIDEr score. This shows that the visual encoder of the AIC model benefits a lot from the global supervision information through joint training with the NAIC decoder.  
As for “w/o. CDC”, which means not performing causal dynamics calibration on any target words at the fine-tuning stage, its performance also decreases, e.g. 0.4 for the BLEU-4 score and 1.5 for the CIDEr score. Moreover, to illustrate the superiority of CDC, we replaced it with the conventional knowledge distillation, i.e., remove the $L_{IA}$ in Equation 14, the results show poor performance.  
These demonstrate the effect of each part in our design as well as incorporating specific future context into the image captioning model on its unconfident words.

\paragraph{\textbf{Effect of Hyper-parameters $\lambda$ and $\epsilon$}}

\begin{figure}
	\begin{center}
		\includegraphics[width=0.9\columnwidth]{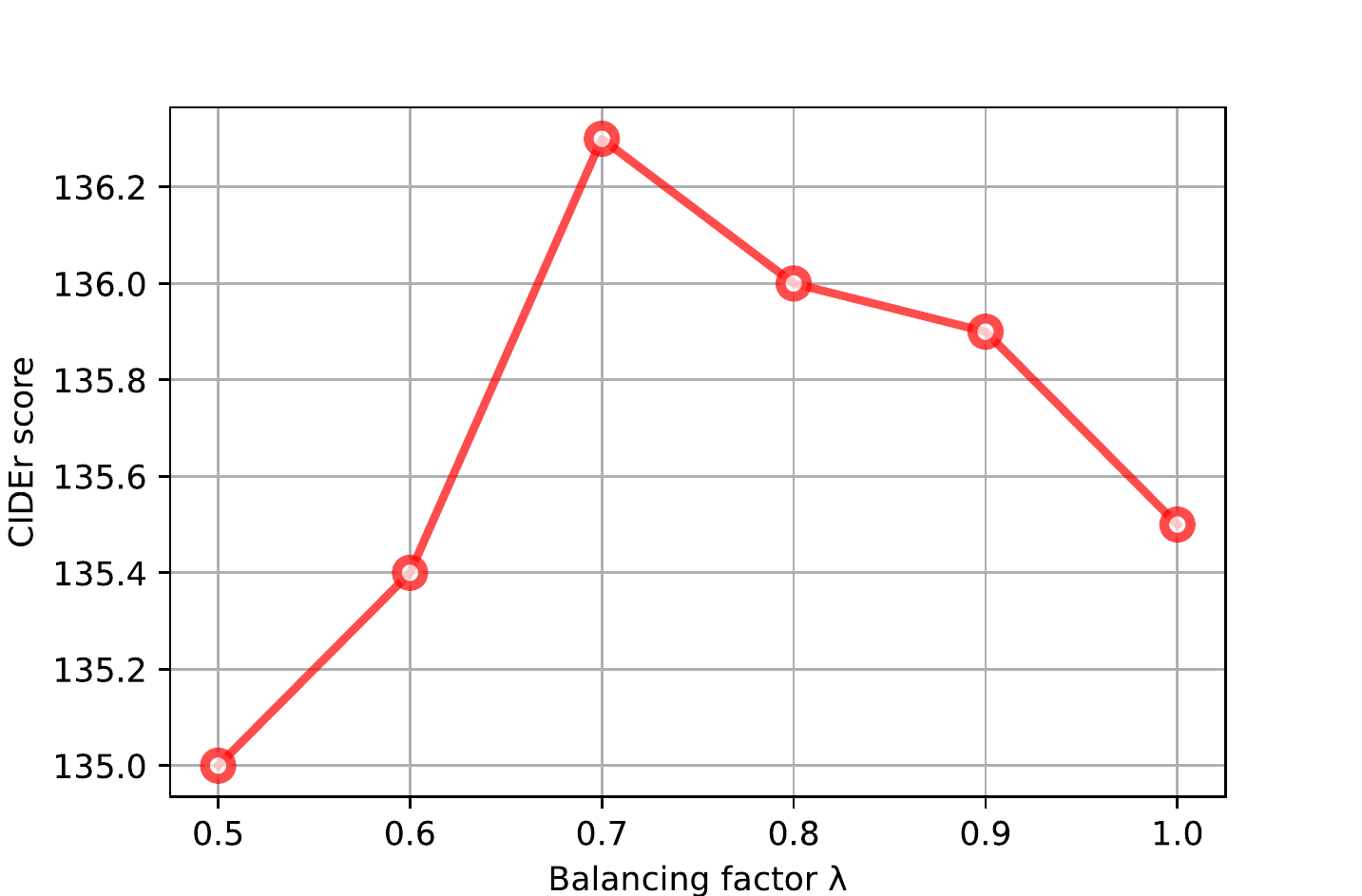}
	\end{center}
	\caption{The evaluated CIDEr scores according to the combine training stage on MS COCO offline test set with different $\lambda$, the balancing factor in different loss.}
	\label{fig:5}
\end{figure}

\begin{figure}
	\begin{center}
		\includegraphics[width=0.9\columnwidth]{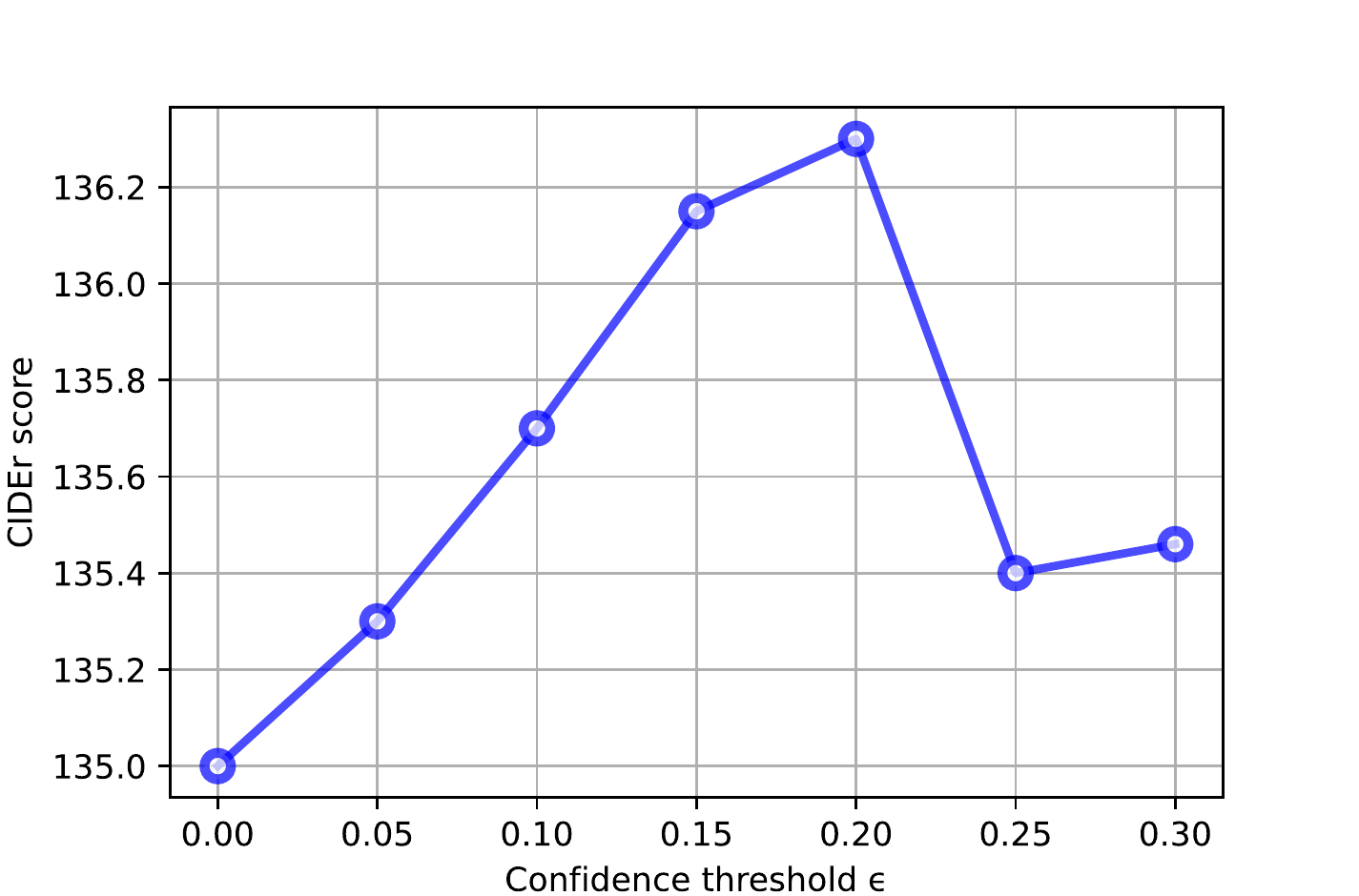}
	\end{center}
	\caption{ The evaluated CIDEr scores on the MS COCO offline test set with different value of $\epsilon$, the confidence threshold for masked words.}
	\label{fig:6}
\end{figure}

There exist important hyper-parameters in the FutureCap framework that we need to tune on the validation set to achieve a good performance, i.e., the balancing factor $\lambda$ in Equation \ref{eq:10} and the confidence threshold $\epsilon$ for determining mask words set $S_m$.
To balance the training of the AIC model and NAIC model at the pre-training stage, we try to select the optimal $\lambda$ that can bring steady improvements to the AIC model. Specifically, we gradually vary $\lambda$ from 0.5 to 1.0 with an increment of 0.1 and evaluate the performance on the validation set, the evaluation results are as shown in Figure \ref{fig:5}. We can see that the final image captioning model achieves its peak when $\lambda = 0.7$. Hence, $\lambda$ is set to 0.7 by default.
Given the selected $\lambda$, at the fine-tuning calibration stage,
we also analyze the impact of $\epsilon$ on the validation set. 
Practically, we change the value of $\epsilon$ from 0.0 to 0.3 with an interval of 0.05. As shown in Figure 6, the AIC model performs the best when the $\epsilon$ comes to 0.2. Therefore, we set $\epsilon$ = 0.2 as the confidence threshold for the causal dynamics calibration stage.

\begin{table}
	\begin{center}
	{\caption{Performance comparisons of incorporating different distribution calibration strategies on the MS COCO test set.}
				\label{tab:4}}
		\setlength{\tabcolsep}{1.5mm}{
			\begin{tabular}{lcccccc}
			\hline
				%	& \multicolumn{5}{|c|}{\textbf{Cross-Entropy Loss}} &\multicolumn{5}{|c|}{\textbf{CIDEr-D Score Optimization}} \\ \cline{2-11}
				Model&B-1 &B-4&M&R&C&S\\
				\hline 
				\hline
				\rowcolor[gray]{.9} FutureCap&\textbf{82.2}&\textbf{40.3}&\textbf{30.1}&\textbf{59.8}&\textbf{136.3}&\textbf{23.8}\\ 
				 Random&81.9&40.2&29.7&59.6&135.8&23.7\\
				 Highest&81.8&40.1&29.6&59.5&135.6&23.6\\
				 Wrong&81.9&40.1&29.7&59.6&135.7&23.6\\
				 OnlyOne&81.7&40.1&29.6&59.5&135.3&23.6\\
			\hline
			\end{tabular}}
	\end{center}
\end{table}

\paragraph{\textbf{Effect of Mask Selection Strategy}}

In our future context modeling framework, for each generated caption pair, we adopt causal dynamics calibration to transfer the knowledge of the NAIC model into the AIC model only on the masked word set $S_m$. 
The set is determined by masking words whose AIC-predicted probabilities of the corresponding ground truths are lower than a pre-set threshold $\epsilon$. It is natural to question if exists other masked word selection patterns and how they perform. Therefore, we further investigate the following four variant masking methods:
\begin{itemize}
    \item \textbf{\emph{Random}}: For the given sentence $S$, randomly select $k$ words to be masked and input to the teacher NAIC model to conduct causal dynamics calibration accordingly.
    \item  \textbf{\emph{Highest}}: As a contrast, we mask the generated words from the AIC model whose predicted probabilities of the ground truth words are higher than the preset threshold $\epsilon$.
    \item \textbf{\emph{Wrong}}: Since the ground-truth labels are given, we try to mask the words where the highest probability predictions of the AIC model are different from the corresponding labels.
    \item \textbf{\emph{OnlyOne}}: In this variant, to illustrate the necessity of selectively distilling knowledge on a portion rather than all of the target words, we generate NAIC-predicted probability distributions for all target words. As an extreme case, we iteratively and only mask one word at once with the given image and residual sentence as input to the NAIC model.
\end{itemize}

The evaluation results for different masked word selection strategies are presented in Table \ref{tab:4}. We can observe that: 1) For “Random” and “Highest” masking strategies, both variants are inferior to our threshold-based causal dynamics calibration method. In particular, the results of “Highest” indicate that conducting dynamics calibration on the confident words is less effective, resulting in a decrease of 0.6 CIDEr score. Meantime, the heuristic selection of masked words is necessary rather than randomness; 2) The result of “Wrong” is lower than our approach. It may be due to the distribution difference between low confidence and incorrect predicted words. 3) “OnlyOne” represents one approach to generating NAIC-predicted probability distributions for all target words iteratively. It is also reasonable for “OnlyOne” to obtain a worse performance since some words can be easy to generate conditioned on local context and over-calibration hold some side effects. 
All these results demonstrate that it is crucial for the AIC model to exploit the global context on its unconfident words. At the same time, a more advanced and learnable selection strategy may contribute to better captioning performance.

% 间隔是0.2还是0.1？ 文和表不一致

\begin{table}
	\begin{center}
	{\caption{The percentage of words within each probability interval on the training set.}
				\label{tab:5}}
		\setlength{\tabcolsep}{0.8mm}{
			\begin{tabular}{lcccccc}
				\hline 
				%	& \multicolumn{5}{|c|}{\textbf{Cross-Entropy Loss}} &\multicolumn{5}{|c|}{\textbf{CIDEr-D Score Optimization}} \\ \cline{2-11}
				Model&[0,0.1) &[0.1,0.2)&[0.2,0.3)&[0.3,0.4)&[0.4,0.5)&[0.5,1)\\
					\hline 
						\hline
				Transformer &26.21&5.13&3.12&4.56&6.14&54.84\\
				FutureCap &25.92&4.78&3.01&4.24&5.92&56.13\\
			\rowcolor[gray]{.9}	$\Delta$ &-0.29&-0.35&-0.11&-0.32&-0.22&+1.29\\
			\hline
			\end{tabular}
		}
	\end{center}
\end{table}

\paragraph{\textbf{Influence on Model Confidence Distribution}}

As the prior experiments show one drawback of conventional image captioning lies in the low confidence to correct words, here we also investigate the change of image captioning model confidence with respect to ground-truth words on the MS COCO training set with future context modeling. We list the percentage of words within each interval, in terms of AIC-predicted probability in  Table \ref{tab:5}. As the probability higher than 0.5 must be the maximum across the vocabulary, we chunk 0.5-1.0 as a high-confidence interval while the others are subdivided into low-confidence intervals. 
According to the evaluation results, it is obvious that the number of words in low-confidence intervals drops with FutureCap. For instance, the number of words located in [0.1, 0.2) becomes 0.35\% fewer. It indicates that our FutureCap model becomes more confident about the ground-truth words given the accurate context.

\subsection{Case Study}

\begin{figure}[t]
	\centering
	\includegraphics[width=0.95\columnwidth]{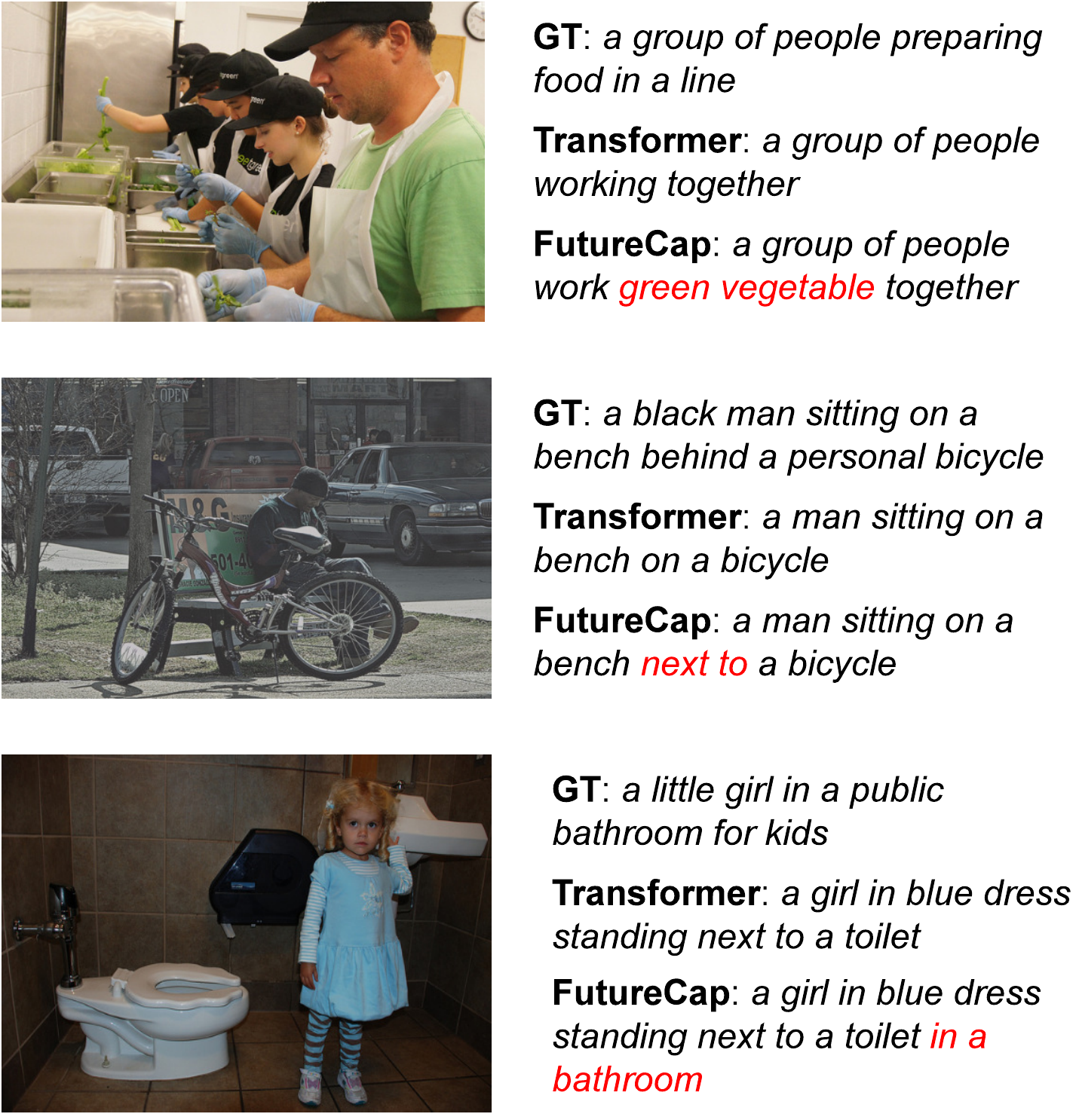}
	\caption{Case studies of original Transformer and our FutureCap model, coupled with the corresponding ground truth sentences (GT).}
	\label{fig:7}
\end{figure}

In order to qualitatively show the effectiveness of future context modeling, we showcase several generated image description results from conventional Transformer with mesh-memory and our FutureCap model, as well as the human-annotated ground-truth sentences (GT) in Figure \ref{fig:7}. Generally, it is easy to see that both approaches are able to produce linguistically coherent descriptions. 
Nevertheless, when examining the fine-grained image content, 
our future information incorporated method produces more accurate and fluent descriptive sentences by exploiting global information for different word predictions. For example, plain Transformer generates the phrase \emph{on a bicycle} which is inconsistent with the visual relationship for the second image, while the words \emph{next to a bicycle} in our model depicts more precisely. This again confirms the advantage of capturing global context when applying the proposed FutureCap method.

\section{Related Works}

\paragraph{\textbf{Image Captioning}}
%\subsection{Image Captioning}
In recent years, a large number of neural systems have been proposed for the image captioning task \cite{Vinyals2015Show, Xu2015Show,Anderson2017Bottom,cornia2020meshed,herdade2019image,huang2019attention,pan2020x,fei2022attention}. The state-of-the-art approaches depend on the encoder-decoder framework to translate the image into a descriptive sentence. Specifically, the
encoder network computes visual representations for the image and the decoder network generates
a target sentence based on the visual representations. To allow more effective use of the visual representations, a series of attention models have been proposed and achieved great success in multiple sequence-to-sequence learning tasks 
\cite{bahdanau2014neural,luong2015effective}.  In recent years, Transformer-based  architectures \cite{li2019entangled,fei2019fast,cornia2020meshed,pan2020x,fei2021memory,yan2021semi,ji2021improving} are introduced to replace conventional RNN, achieving new state-of-the-art performances.
On the other hand, lots of mask-based non-autoregressive decoding methods are studied for inference acceleration with a global perspective \cite{fei2019fast,guo2020non,gao2019masked,fei2020iterative,fei2021partially}. 
However, as far as we are concerned, improving the original language decoding with supervised future information from the NAIC decoder has never been studied in image captioning, which pushes forward our exploration in this paper.

\paragraph{\textbf{Training Procedure}}
%\subsection{Training Procedure}
Training strategy for image captioning models usually follows the word-level cross-entropy paradigm from left to right. This was later combined with a fine-tuning phase based on the application of the REINFORCE method, to allow use as optimization objectives captioning
metrics directly \cite{Rennie2017Self, liu2017improved}, boosting the final performance. 
As a strategy to improve both training phases, in \cite{huang2020teacher} it is proposed to exploit a teacher model trained on image attributes to generate additional supervision signals for the captioning model. These are in the form of soft labels, which the captioning model has to align within the cross-entropy phase, and re-weighting of the caption words to guide the fine-tuning phase. \cite{barraco2022camel} improve the quality with the interaction of two interconnected language models that learn from each other. 
Additional improvement to the performance of recent self-attention-based image captioning approaches is due to the use of large-scale vision-and-language pre-training \cite{cornia2021universal, li2020oscar,zhang2021vinvl,zhou2020unified,radford2021learning}, which can be done on noisy and weakly annotated image-text pairs, also exploiting pre-training losses different from cross-entropy, such as the masked word loss \cite{zhang2021vinvl}. Different from all previous methods, our approach is based on the assistance of an additional non-autoregressive image captioning model that is trained with the muti-task learning and dynamic distribution calibration, without changing the internal model architecture or relying on a prior large-scale pre-training model.

\paragraph{\textbf{Future Information Incorporation}}
%\subsection{Future Information Incorporation}
There are numerous works \cite{chen2020distilling,kafle2016answer, duan2021modeling,qin2019look,ren2017deep,ma2020learning,fei2020actor} dived to exploit the future information to boost the performance for sequence-to-sequence learning. However, their modelings are different from ours. Specifically, to exploit the future information, \cite{chen2020distilling} adopt a fine-tuned BERT \cite{devlin2018bert} to encode the words that will be generated in the future to acquire the global cost and then exploited as extra supervision to guide the current word generation. 
\cite{zhou2022confidence,ai2021almost} employ an extra teacher network to help the neural machine translation model capture global information with knowledge distillation.  
For \cite{duan2021modeling, qin2019look,ren2017deep}, given the previous history, in addition to the current target, they further predict the future words, \emph{i.e.}, \cite{duan2021modeling, qin2019look} one more step ahead, and \cite{ren2017deep} the rest of the sequence.
\cite{ma2020learning} only consider the current target to model the future information and \cite{wang2018image,sammani2020show,stefanini2021show} regularize the right-to-left generation, while we directly leverage the effective knowledge to enhance the modeling of the future information. The most similar work is \cite{zhou2022compact}, both works point the importance of bi-direction context and employ it for improved image captioning. In contrast, \cite{zhou2022compact} introduce a compact directional transformer to parallel decoding while we devise causal dynamics calibration without extra parameters.

\section{Conclusion}

In this paper, we focus on making a conventional image captioning model to effectively exploit the global context without any extra inference cost. Specifically, we resort to the mask-based non-autoregressive decoder for future information modeling during training. Specifically, we introduce multi-task learning to benefit the AIC model by sharing its visual encoder with an auxiliary NAIC. Next, we explore distilling a teacher NAIC model by training the AIC student model to capture the causal dynamics for unconfident words. Experimental results on the MS COCO dataset show that our future information incorporation framework can significantly improve the captioning performance. More importantly, no additional carefully designed network is needed and only the original image captioning model is involved during inference.


% \newpage 
\bibliographystyle{ACM-Reference-Format}
\bibliography{sample-base}

\end{document}